\documentclass[10pt,letterpaper]{article}
\usepackage[top=0.85in,left=2.75in,footskip=0.75in]{geometry}

\usepackage{changepage}

\usepackage[utf8]{inputenc}

\usepackage{textcomp,marvosym}

\usepackage{fixltx2e}

\usepackage{amsmath,amssymb}

\usepackage{cite}

\usepackage{nameref,hyperref}

\usepackage[right]{lineno}

\usepackage{microtype}
\DisableLigatures[f]{encoding = *, family = * }

\usepackage{rotating}

\raggedright
\usepackage[aboveskip=1pt,labelfont=bf,labelsep=period,justification=raggedright,singlelinecheck=off]{caption}

\makeatletter
\renewcommand{\@biblabel}[1]{\quad#1.}
\makeatother

\date{}

\usepackage{lastpage,fancyhdr,graphicx}
\usepackage{epstopdf}
\fancyheadoffset[L]{2.25in}
\fancyfootoffset[L]{2.25in}
\usepackage{bm,cite,array,fixltx2e,url}

\newcommand{\BigO}[1]{\ensuremath{\operatorname{O}\bigl(#1\bigr)}}
\bmdefine\bmu{\mu}
\bmdefine\bsigma{\sigma}
\bmdefine\bLambda{\Lambda}
\newcommand{\tab}{\hspace*{0.9em}}
\graphicspath{{img/}}
\DeclareGraphicsExtensions{.pdf,.jpeg,.png}

\usepackage{algorithm}
\usepackage{algorithmic}

\begin{document}
\vspace*{0.35in}

\begin{flushleft}
{\Large
\textbf\newline{A Fast Incremental Gaussian Mixture Model - Submission to PLOS Journals}
}
\newline
\\
Rafael Pinto\textsuperscript{1},
Paulo Engel\textsuperscript{1},
\\
\bigskip
\bf{1} Instituto de Informática, Universidade Federal do Rio Grande do Sul, Porto Alegre, RS, Brazil
\bigskip

Av. Bento Gonçalves, 9500 - Agronomia - Porto Alegre, RS - Zip 91501-970 - Brazil

\{rcpinto,engel\}@inf.ufrgs.br

\end{flushleft}
\section*{Abstract}
This work builds upon previous efforts in online incremental learning, namely the Incremental Gaussian Mixture Network (IGMN). The IGMN is capable of learning from data streams in a single-pass by improving its model after analyzing each data point and discarding it thereafter. Nevertheless, it suffers from the scalability point-of-view, due to its asymptotic time complexity of \BigO{NKD^3} for $N$ data points, $K$ Gaussian components and $D$ dimensions, rendering it inadequate for high-dimensional data. In this paper, we manage to reduce this complexity to \BigO{NKD^2} by deriving formulas for working directly with precision matrices instead of covariance matrices. 
The final result is a much faster and scalable algorithm which can be applied to high dimensional tasks. This is confirmed by applying the modified algorithm to high-dimensional classification datasets.

\section{Introduction}
	The Incremental Gaussian Mixture Network (IGMN) \cite{heinen2012using,heinen2011igmn} is a supervised algorithm which approximates the EM algorithm \cite{engel2011incremental}. It	
	creates and continually adjusts a probabilistic model consistent to all sequentially presented data, after each data point 
	presentation, and without the need to store any past data points. Its learning process is aggressive,  meaning 
	that only a single scan through the data is necessary to obtain a consistent model.
	
	IGMN adopts a Gaussian mixture model of distribution components that can be 
	expanded to accommodate new information from an input data point, or reduced if spurious components are identified along the 
	learning process. Each data point assimilated by the model contributes to the sequential update of the model parameters based 
	on the maximization of the likelihood of the data. The parameters are updated through the accumulation of relevant 
	information extracted from each data point.
	
	The IGMN is capable of supervised learning, simply by assigning any of its input 
	vector elements as outputs (any element can be used to predict any other element, like autoassociative neural networks \cite{rumelhart1986parallel}). This feature is useful for simultaneous learning of forward and inverse kinematics, as well as for simultaneous learning of a value function and a policy in reinforcement learning \cite{heinen2011igmnRL}. 
	
	However, the IGMN suffers from cubic time complexity due to matrix inversion operations and determinant computations. Its time complexity is of \BigO{NKD^3}, where $N$ is the number of data points, $K$ is the number of Gaussian components and $D$ is the problem dimension. It makes the algorithm prohibitive for high-dimensional tasks (like visual tasks) and thus of limited use. One solution would be to use diagonal covariance matrices, but this decreases the quality of the results, as already reported in previous work \cite{heinen2011connectionist,pinto2011echo}. In this work, we propose the use of rank-one updates for both inverse matrices and determinants applied to full covariance matrices, thus reducing the time complexity to \BigO{NKD^2} for learning while keeping the quality of a full covariance matrix solution.
	
	For the specific case of the IGMN algorithm, to the best of our knowledge, this has not been tried before, although we can find similar efforts for related algorithms. In \cite{Salmen20101903}, rank-one updates were applied to an iterated linear discriminant analysis algorithm in order to decrease the complexity of the algorithm. Rank-one updates were also used in \cite{lefakis2014jointly}, where Gaussian models are employed for feature selection. Finally, in	\cite{olsen2001extended}, the same kind of optimization was applied to Maximum Likelihood Linear Transforms (MLLT).
	
	The next Section describes the algorithm in more detail with the latest improvements to date. Section \ref{sec:figmn} describes our improvements to the algorithm. Section \ref{sec:experiments} shows the experiments and results obtained from both versions of the IGMN for comparison, and Section \ref{sec:conclusion} finishes this work with concluding remarks.



\section{Incremental Gaussian Mixture Network}\label{sec:igmn}

In the next subsections we describe the current version of the IGMN algorithm. 
	

\subsection{Learning} \label{sec:learning}
				
	The algorithm starts with no components, which are created as necessary (see subsection \ref{sec:create}). Given input 
	$\textbf{x}$ (a single instantaneous data point), the IGMN algorithm processing step is as follows. First, the squared Mahalanobis distance $d^2(\textbf{x},j)$ for each component $j$ is computed:
	
	\begin{equation}\label{equ:igmn-maha}
		d^2_M(\textbf{x},j) =  (\textbf{x}-\bmu_j)^T \textbf{C}^{-1}_j (\textbf{x}-\bmu_j)
	\end{equation}

	
	where 
	$\bmu_j$ is the $j^{th}$ component mean, $\textbf{C}_j$ its full covariance 
	matrix 
	. If any $d^2(\textbf{x},j)$ is smaller than than $\chi^2_{D,1-\beta}$ (the $1-\beta$ percentile of a chi-squared distribution with $D$ degrees-of-freedom, where $D$ is the input dimensionality and $\beta$ is a user defined meta-parameter, e.g., $0.1$)
	, an update will occur, and posterior probabilities are calculated for each component as follows:

	\begin{equation}\label{equ:igmn-like}
		p(\textbf{x}|j) = \frac{1}{(2\pi)^{D/2}\sqrt{|\textbf{C}_j|}} \exp\left(-\frac{1}{2} d^2_M(\textbf{x},j) \right)
	\end{equation}

	\begin{equation}\label{equ:posterior}
		p(j|\textbf{x}) = \frac{ p(\textbf{x}|j) p(j) }{ \displaystyle\sum\limits_{k=1}^K{ 
		 p(\textbf{x}|k) p(k) } }	\tab \forall{j}
	\end{equation}	

	where $K$ is the number of components. Now, parameters of the algorithm must be updated according to the following equations:
	
	\begin{equation}\label{equ:igmn-v}
		v_j(t) = v_j(t-1) + 1
	\end{equation}

	\begin{equation}\label{equ:igmn-sp}
		sp_j(t) = sp_j(t-1) + p(j|\textbf{x})
	\end{equation}
	
	\begin{equation}\label{equ:igmn-e}
		\textbf{e}_j = \textbf{x} - \bmu_j(t-1)
	\end{equation}
	
	\begin{equation}\label{equ:igmn-omega}
		\omega_j = \frac{ p(j|\textbf{x}) } { sp_j }
	\end{equation}

	\begin{equation}\label{equ:igmn-delta}
		\Delta\bmu_j = \omega_j \textbf{e}_j
	\end{equation}

	\begin{equation}\label{equ:igmn-mu}
		\bmu_j(t) = \bmu_j(t-1) + \Delta\bmu_j
	\end{equation}
	
	\begin{equation}\label{equ:igmn-enew}
		\textbf{e}^*_j = \textbf{x} - \bmu_j(t)
	\end{equation}

	\begin{equation}\label{equ:igmn-C}
		\textbf{C}_j(t) = (1-\omega_j) \textbf{C}_j(t-1) + \omega_j \textbf{e}_j^* \textbf{e}_j^{*T} - \Delta\bmu_j \Delta\bmu_j^{T}
	\end{equation}

	\begin{equation} \label{equ:igmn-p}
		p(j) = \frac{ sp_j } { \displaystyle\sum\limits_{q=1}^M{sp_q} }
	\end{equation}
	
	where $sp_j$ and $v_j$ are the accumulator and the age of component $j$, respectively, and $p(j)$ is its prior probability.

	
\subsection{Creating New Components} \label{sec:create}
	
	If the update condition in the previous subsection is not met, then a new component $j$ is created and initialized as follows:
			
	\[
		\bmu_j = \textbf{x}; \tab
		sp_j = 1; \tab
		v_j = 1; \tab
		p(j) = \frac{ 1 } { \displaystyle\sum\limits_{i=1}^K{sp_i} }; \tab
		\textbf{C}_j = \bsigma_{ini}^2 \textbf{I}
	\]
	where $K$ already includes the new component and $\bsigma_{ini}$ can be obtained by:	
	\begin{equation}\label{equ:sigma} 
		\bsigma_{ini} = \delta std(\textbf{x})
	\end{equation}
	
	where $\delta$ is a manually chosen scaling factor (e.g., 0.01) and $std$ is the standard deviation of the dataset. Note that the IGMN is an online and incremental algorithm and therefore it may be the case that we do not have the entire dataset to extract descriptive statistics. In this case the standard deviation can be just an estimation (e.g., based on sensor limits from a robotic platform), without impacting the algorithm.

\subsection{Removing Spurious Components} \label{sec:removing}
	A component $j$ is removed whenever $v_j > v_{min}$ and $sp_j < sp_{min}$, where $v_{min}$ and $sp_{min}$ are manually chosen (e.g., 5.0 and 3.0,
	respectively). In that case, also, $p(k)$ must be adjusted for all $k \in K$, $k \ne j$, using (\ref{equ:igmn-p}). In other words, each component is given some time $v_{min}$ to show its importance to the model in the form of an accumulation of its posterior probabilities $sp_j$.

\subsection{Inference} \label{sec:recalling}


	In the IGMN, any element can be predicted by any other element. This is done by reconstructing data from the target elements
	($\textbf{x}_t$, a slice of the entire input vector $\textbf{x}$) by estimating the posterior probabilities using only the given elements ($\textbf{x}_i$, also a slice of the entire input vector $\textbf{x}$), as follows:

	\begin{equation}\label{equ:recall}
		p(j|\textbf{x}_i) = \frac{ p(\textbf{x}_i|j) p(j) }{ \displaystyle\sum\limits_{q=1}^M{ 
		 p(\textbf{x}_i|q) p(q) } }	\tab \forall{j}
	\end{equation}		
	It is similar to (\ref{equ:posterior}), except that it uses a modified input vector $\textbf{x}_i$ with the target 
	elements $\textbf{x}_t$ removed from calculations. After that, $\textbf{x}_t$ can be reconstructed using the conditional mean equation:
	
	\begin{equation}\label{equ:reconstructfull}
		\hat{\textbf{x}_t} = \displaystyle\sum\limits_{j=1}^M{ p(j|\textbf{x}_i) (\bmu_{j,t} + \textbf{C}_{j,ti} \textbf{C}_{j,i}^{-1} (\textbf{x}_i - \bmu_{j,i})) }
	\end{equation}	

	where $\textbf{C}_{j,ti}$ is the submatrix of the $j$th component covariance matrix associating the unknown and known 
	parts of the data, $\textbf{C}_{j,i}$ is the submatrix corresponding to the known part only and $\bmu_{j,i}$ is the $j$th's 
	component mean without the element corresponding to the target element. 

\section{Fast IGMN}\label{sec:figmn}
One of the contributions of this work lies in the fact that Equation \ref{equ:igmn-maha} (the squared Mahalanobis distance) requires a matrix inversion, which has a asymptotic time complexity of \BigO{D^3}, for $D$ dimensions (\BigO{D^{log_2 7 + \BigO{1}}} for the Strassen algorithm or at best \BigO{D^{2.3728639}} with the most recent algorithms to date \cite{gall2014powers}). This renders the entire IGMN algorithm as impractical for high-dimension tasks. Here we show how to work directly with the inverse of covariance matrix (also called the precision or concentration matrix) for the entire procedure, therefore avoiding costly inversions.

Firstly, let us denote $\textbf{C}^{-1} = \bLambda$, the precision matrix. Our task is to adapt all equations involving $\textbf{C}$ to instead use $\bLambda$. 

We now proceed to adapt Equation \ref{equ:igmn-C} (covariance matrix update). This equation can be seen as a sequence of two rank-one updates to the $\textbf{C}$ matrix, as follows:

	\begin{equation}\label{equ:figmn-C1}
		\bar{\textbf{C}}_j(t) = (1-\omega_j) \textbf{C}_j(t-1) + \omega_j \textbf{e}_j^* \textbf{e}_j^{*T}
	\end{equation}

	\begin{equation}\label{equ:figmn-C2}
		\textbf{C}_j(t) = \bar{\textbf{C}}_j(t) - \Delta\bmu_j \Delta\bmu_j^{T}
	\end{equation}

This allows us to apply the Sherman-Morrison formula \cite{sherman1950}:

\begin{equation}\label{equ:sherman}
(\textbf{A} + \textbf{uv}^T)^{-1} = \textbf{A}^{-1} - \frac{\textbf{A}^{-1} \textbf{uv}^T \textbf{A}^{-1}}{1 + \textbf{v}^T \textbf{A}^{-1} \textbf{u}}
\end{equation}

This formula shows how to update the inverse of a matrix plus a rank-one update. For the second update, which subtracts, the formula becomes:

\begin{equation}\label{equ:sherman2}
(\textbf{A} - \textbf{uv}^T)^{-1} = \textbf{A}^{-1} + \frac{\textbf{A}^{-1} \textbf{uv}^T \textbf{A}^{-1}}{1 - \textbf{v}^T \textbf{A}^{-1} \textbf{u}}
\end{equation}

In the context of IGMN, we have $\textbf{A} = (1-\omega) \textbf{C}_j(t-1) = (1-\omega) \bLambda_j^{-1}(t-1)$ and $\textbf{u} = \textbf{v} = \sqrt{\omega}\textbf{e}^*$ for the first update, while for the second one we have $\textbf{A} = \bar{\textbf{C}}_j(t)$ and $\textbf{u} = \textbf{v} = \Delta\bmu_j$. Rewriting \ref{equ:sherman} and \ref{equ:sherman2} we get (for the sake of compactness, assume all subscripts for $\bLambda$ and $\Delta\bmu$ to be $j$):

\begin{equation}\label{equ:figmn-sherman1}
\bar{\bLambda}(t) = \frac{\bLambda(t-1)}{1-\omega} - \frac{\frac{\omega}{(1-\omega)^2}\bLambda(t-1) \textbf{e}^*\textbf{e}^{*T} \bLambda(t-1)}{1 + \frac{\omega}{1-\omega}\textbf{e}^{*T} \bLambda(t-1) \textbf{e}^*}
\end{equation}

\begin{equation}\label{equ:figmn-sherman2}
\bLambda(t) = \bar{\bLambda}(t) + \frac{\bar{\bLambda}(t) \Delta\bmu\Delta\bmu^T \bar{\bLambda}(t)}{1 - \Delta\bmu^T \bar{\bLambda}(t) \Delta\bmu}
\end{equation}

These two equations allow us to update the precision matrix directly, eliminating the need for the covariance matrix $\textbf{C}$. They have $\BigO{N^2}$ complexity due to matrix-vector products.

Following on the adaptation of the IGMN equations, Equation \ref{equ:igmn-maha} (the squared Mahalanobis distance) allows for a direct substituion, yielding the following new equation:

	\begin{equation}\label{equ:figmn-maha}
		d^2_M(\textbf{x},j) =  (\textbf{x}-\bmu_j)^T \bLambda_j (\textbf{x} -\bmu_j)
	\end{equation}
	
	which now has a $\BigO{N^2}$ complexity, since there is no matrix inversion as the original equation. After removing the cubic complexity from this step, the determinant computation will be dealt with next.

Since the determinant of the inverse of a matrix is simply the inverse of the determinant, it is sufficient to invert the result. But computing the determinant itself is also a \BigO{D^3} operation, so we will instead perform rank-one updates using the Matrix Determinant Lemma \cite{harville2008matrix}, which states the following:

	\begin{equation}\label{equ:figmn-det-lemma}
	    |\textbf{A} + \textbf{u}\textbf{v}^T| = |\textbf{A}| (1 + \textbf{v}^T \textbf{A}^{-1} \textbf{u})
	\end{equation}
	\begin{equation}\label{equ:figmn-det-lemma2}
	    |\textbf{A} - \textbf{u}\textbf{v}^T| = |\textbf{A}| (1 - \textbf{v}^T \textbf{A}^{-1} \textbf{u})
	\end{equation}

Since the IGMN covariance matrix update involves a rank-two update, adding a term and then subtracting one, both rules must be applied in sequence, similar to what has been done with the $\bLambda$ equations. Equations \ref{equ:figmn-C1} and \ref{equ:figmn-C2} may be reused here, together with the same substitutions previously showed, leaving us with the following new equations for updating the determinant (again, $j$ subscripts were dropped):

	\begin{equation}\label{equ:figmn-det-lemma-apply1}
	    |\bar{\textbf{C}}(t)| = (1-\omega)^D |\textbf{C}(t-1)| \left(1 + \frac{\omega}{1 - \omega}\textbf{e}^{*T} \bLambda(t-1) \textbf{e}^*\right)
	\end{equation}

	\begin{equation}\label{equ:figmn-det-lemma-apply2}
	    |\textbf{C}(t)| = |\bar{\textbf{C}}(t)| (1 - \Delta\bmu^{T} \bar{\bLambda}(t) \Delta\bmu)
	\end{equation}

This was the last source of cubic complexity, which is now quadratic. 

Finishing the adaptation in the learning part of the algorithm, we just need to define the initialization for $\bLambda$ for each component. What previously was $\textbf{C}_j = \bsigma^2_{ini}\textbf{I}$ now becomes $\bLambda_j = \bsigma^{-2}_{ini}\textbf{I}$, the inverse of the variances of the dataset. Since this matrix is diagonal, there are no costly inversions involved. And for initializing the determinant $|\textbf{C}|$, just set it to $\prod{\bsigma_{ini}^2}$, which again takes advantage of the initial diagonal matrix to avoid costly operations. Note that we keep the precision matrix $\bLambda$, but the determinant of the covariance matrix $\textbf{C}$ instead. See algorithms \ref{alg:figmn-learn} to \ref{alg:figmn-create} for a summary of the new learning algorithm (excluding pruning, for brevity).

\begin{algorithm}[ht]
\begin{algorithmic}
\caption{Fast IGMN Learning}\label{alg:figmn-learn}
\REQUIRE{$\delta$,$\beta$,$\textbf{X}$}
\STATE $K > 0$, $\bsigma^{-1}_{ini} = (\delta std(\textbf{X}))^{-1}, M = \emptyset$
\FORALL{input data vector $\textbf{x} \in \textbf{X}$}
    \IF {$K = 0$ \OR $\exists j$, $d^2_M(\textbf{x},j) < \chi^2_{D,1-\beta}$}
        \STATE $update(\textbf{x})$
    \ELSE
        \STATE $M \gets M \cup create(\textbf{x})$
    \ENDIF
\ENDFOR
\end{algorithmic}
\end{algorithm}

\begin{algorithm}[ht]
\begin{algorithmic}
\caption{update}\label{alg:figmn-update}
\REQUIRE{$\textbf{x}$}
\FORALL{Gaussian component $j \in M$}
\STATE Compute equations \ref{equ:igmn-maha} to \ref{equ:igmn-p} substituting \ref{equ:igmn-maha} for \ref{equ:figmn-maha} and \ref{equ:igmn-C} for \ref{equ:figmn-sherman1} and \ref{equ:figmn-sherman2}
\STATE Compute equations \ref{equ:figmn-det-lemma-apply1} and \ref{equ:figmn-det-lemma-apply2}

\ENDFOR
\end{algorithmic}
\end{algorithm}

\begin{algorithm}[ht]
\begin{algorithmic}
\caption{create}\label{alg:figmn-create}
\REQUIRE{$\textbf{x}$}
\STATE $K \gets K + 1$
\RETURN new Gaussian component $K$ with $\bmu_K = \textbf{x}$, $\bLambda_K = \bsigma^{-1}_{ini}\textbf{I}$, $|\textbf{C}_K| = |\bLambda_K|^{-1}$, $sp_j = 1$, $v_j = 1$, $p(j) = \frac{1}{\displaystyle\sum\limits_{k=1}^K{sp_i}}$
\end{algorithmic}
\end{algorithm}

Finally, the inference Equation \ref{equ:reconstructfull} must also be updated in order to allow the IGMN to work in supervised mode. This can be accomplished by the use of a block matrix decomposition (note that here $\textbf{C}$ is just another sub-matrix, not the covariance matrix as used before):

$
\bLambda = \begin{bmatrix}
\textbf{A} & \textbf{B} \\
\textbf{C} & \textbf{D}
\end{bmatrix}^{-1} = 
\begin{bmatrix}
\textbf{X} & \textbf{Y} \\
\textbf{Z} & \textbf{W}
\end{bmatrix} = 
\begin{bmatrix}
(\textbf{A} - \textbf{BD}^{-1}\textbf{C})^{-1} & -\textbf{A}^{-1}\textbf{B}(\textbf{D} - \textbf{CA}^{-1}\textbf{B})^{-1} \\                 -\textbf{D}^{-1}\textbf{C}(\textbf{A} - \textbf{BD}^{-1}\textbf{C})^{-1} & (\textbf{D} - \textbf{CA}^{-1}\textbf{B})^{-1}  
\end{bmatrix}  
$

Here, according to Equation \ref{equ:reconstructfull}, we need $\textbf{C}$ and $\textbf{A}^{-1}$. But since the terms that constitute these sub-matrices are relative to the original covariance matrix (which we do not have), they must be extracted from the precision matrix directly. Looking at the decomposition, it is clear that $\textbf{YW}^{-1} = -\textbf{A}^{-1}\textbf{B} = -\textbf{C}\textbf{A}^{-1}$ (the terms between parenthesis in $\textbf{Y}$ and $\textbf{W}$ cancel each other, while $\textbf{B} = \textbf{C}^T$ due to symmetry). So Equation \ref{equ:reconstructfull} can be rewritten as:

	\begin{equation}\label{equ:figmn-reconstructfull}
		\hat{\textbf{x}_t} = \displaystyle\sum\limits_{j=1}^M{ p(j|\textbf{x}_i) (\bmu_{j,t} - \textbf{YW}^{-1} (\textbf{x}_i - \bmu_{j,i})) }
	\end{equation}	

where $\textbf{Y}$ and $\textbf{W}$ can be extracted directly from $\bLambda$. However, we still need to compute the inverse of $\textbf{W}$. 
So we can say that this particular implementation has $\BigO{NKD^2}$ complexity for learning and $\BigO{NKD^3}$ for inference. The reason for us to not worry about that is that $d = i + o$, where $i$ is the number of inputs and $o$ is the number of outputs. The inverse computation acts only upon the output portion of the matrix. Since, in general, $o \ll i$ (in many cases even $o = 1$), the impact is minimal, and the same applies to the $\textbf{YW}^{-1}$ product. In fact, Weka (the data mining platform used in this work \cite{hall2009weka}) allows for only 1 output, leaving us with just scalar operations.

\section{Experiments}\label{sec:experiments}

In order to evaluate the performance of the proposed algorithm, 11 classification tasks (Table \ref{datasets}) were given to the original and improved IGMN algorithms ($\delta = 1$ and $\beta = 0$, so a single component was created for each run and we could focus on speed ups only due to dimensionality). Results were obtained from 2-fold cross-validation and statistical significances came from paired t-tests with $p = 0.05$.

\begin{table}[thb]
\caption{\label{datasets}Datasets}
\scriptsize
{\centering \begin{tabular}{lccc}
\\
\hline
Dataset & Instances (N) & Attributes (D) & Classes \\
\hline
breast-cancer & 286 & 9 & 2 \\
german-credit & 1000 & 20 & 2 \\
pima-diabetes & 768 & 8 & 2 \\
Glass & 214 & 9 & 7 \\
ionosphere & 351 & 34 & 2 \\
iris & 150 & 4 & 3 \\
labor-neg-data & 57 & 16 & 2 \\
soybean & 683 & 35 & 19 \\
twospirals & 193 & 2 & 2 \\
MNIST \cite{lecun1998gradient} (subset) & 1000 & 784 & 10 \\
CIFAR-10 \cite{krizhevsky2009learning} (subset) & 1000 & 3072 & 10 \\
CIFAR-10b (subset) & 100 & 3072 & 10 \\
\hline
\end{tabular} \scriptsize \par}
\end{table}


This experiment was meant to verify that both IGMN implementations produce exactly the same results, which was confirmed, as well as to show that the proposed improvements really deliver the expected speed up (the Weka packages for both variations of the IGMN algorithm can be found at \url{http://www.inf.ufrgs.br/~rcpinto}). This was also confirmed, as can be seen in tables \ref{trainingtime} and \ref{testingtime} (note that the experiments were divided into training and test phases just for comparison purposes, but IGMN is in fact an online algorithm; also note that standard deviations were rounded to 3 decimal places and, in fact, there are not any null standard deviations in the results). Our improved algorithm could deliver a speed up of 2 orders of magnitude in training time (learning) for the CIFAR-10 subset, which follows our expectations according to the new time complexity. Datasets with less dimensions could benefit from the improvements too, albeit in a smaller scale, which was also expected. The other confirmation came from the testing times (inference): they were also improved, since the inverse matrix computation was eliminated from the probability density equation, which is also necessary for inference, and the matrix multiplications and inversions are actually scalar operations, having no impact. In fact, the speed up for inference was around 3 orders of magnitude for the CIFAR-10 subset.

\begin{table}[thb]
\caption{\label{trainingtime}Training Time (in seconds)}
\scriptsize
{\centering \begin{tabular}{lr@{\hspace{0cm}}c@{\hspace{0cm}}rr@{\hspace{0cm}}c@{\hspace{0cm}}r@{\hspace{0.1cm}}c}
\\
\hline
Dataset & \multicolumn{3}{c}{IGMN}& \multicolumn{4}{c}{Fast IGMN} \\
\hline
breast-cancer &     0.010 & $\pm$ &    0.004 &   0.006 & $\pm$ & 0.000 &          \\
german-credit &     0.031 & $\pm$ &    0.012 &   0.016 & $\pm$ & 0.000 &          \\
pima-diabetes &     0.013 & $\pm$ &    0.000 &   0.010 & $\pm$ & 0.001 &          \\
Glass &     0.008 & $\pm$ &    0.000 &   0.005 & $\pm$ & 0.000 & $\bullet$\\
ionosphere &     0.022 & $\pm$ &    0.002 &   0.014 & $\pm$ & 0.002 & $\bullet$\\
iris &     0.005 & $\pm$ &    0.000 &   0.007 & $\pm$ & 0.001 &          \\
labor-neg-data &     0.006 & $\pm$ &    0.001 &   0.007 & $\pm$ & 0.001 &          \\
soybean &     0.042 & $\pm$ &    0.004 &   0.030 & $\pm$ & 0.003 &          \\
twospirals &     0.005 & $\pm$ &    0.001 &   0.006 & $\pm$ & 0.001 &   $\circ$\\
MNIST &   281.257 & $\pm$ &    3.157 &  10.675 & $\pm$ & 0.272 & $\bullet$\\
CIFAR-10 & 20768.494 & $\pm$ & 1244.221 & 175.243 & $\pm$ & 1.190 & $\bullet$\\
\hline
Average &  1754.417 &       &          &  15.526 &       &       &          \\
\hline
\multicolumn{6}{c}{$\circ$, $\bullet$ statistically significant increase or decrease in time}\\
\end{tabular} \scriptsize \par}
\end{table}

\begin{table}[thb]
\caption{\label{testingtime}Testing Time (in seconds)}
\scriptsize
{\centering \begin{tabular}{lr@{\hspace{0cm}}c@{\hspace{0cm}}rr@{\hspace{0cm}}c@{\hspace{0cm}}r@{\hspace{0.1cm}}c}
\\
\hline
Dataset & \multicolumn{3}{c}{IGMN}& \multicolumn{4}{c}{Fast IGMN} \\
\hline
breast-cancer &     0.001 & $\pm$ &    0.000 & 0.001 & $\pm$ & 0.001 &          \\
german-credit &     0.013 & $\pm$ &    0.004 & 0.002 & $\pm$ & 0.000 &          \\
pima-diabetes &     0.004 & $\pm$ &    0.000 & 0.001 & $\pm$ & 0.000 & $\bullet$\\
Glass &     0.001 & $\pm$ &    0.000 & 0.001 & $\pm$ & 0.000 &          \\
ionosphere &     0.010 & $\pm$ &    0.001 & 0.008 & $\pm$ & 0.008 &          \\
iris &     0.000 & $\pm$ &    0.000 & 0.001 & $\pm$ & 0.000 &   $\circ$\\
labor-neg-data &     0.000 & $\pm$ &    0.000 & 0.001 & $\pm$ & 0.001 &          \\
soybean &     0.026 & $\pm$ &    0.007 & 0.004 & $\pm$ & 0.000 &          \\
twospirals &     0.000 & $\pm$ &    0.000 & 0.000 & $\pm$ & 0.000 &          \\
MNIST &   225.262 & $\pm$ &    5.638 & 0.607 & $\pm$ & 0.149 & $\bullet$\\
CIFAR-10 & 17821.407 & $\pm$ & 1699.599 & 7.793 & $\pm$ & 0.098 &          \\
\hline
Average &  1504.097 &       &          & 0.702 &       &       &          \\
\hline
\multicolumn{6}{c}{$\circ$, $\bullet$ statistically significant increase or decrease in time}\\
\end{tabular} \scriptsize \par}
\end{table}

\begin{table}[thb]
\caption{\label{accuracy}Area Under ROC Curve}
\scriptsize
{\centering \begin{tabular}{lr@{\hspace{0cm}}c@{\hspace{0cm}}rr@{\hspace{0cm}}c@{\hspace{0cm}}r@{\hspace{0.1cm}}cr@{\hspace{0cm}}c@{\hspace{0cm}}r@{\hspace{0.1cm}}cr@{\hspace{0cm}}c@{\hspace{0cm}}r@{\hspace{0.1cm}}cr@{\hspace{0cm}}c@{\hspace{0cm}}r@{\hspace{0.1cm}}cr@{\hspace{0cm}}c@{\hspace{0cm}}r@{\hspace{0.1cm}}c}
\\
\hline
Dataset & \multicolumn{3}{c}{Neural Network}& \multicolumn{4}{c}{1-NN} & \multicolumn{4}{c}{Naive Bayes} & \multicolumn{4}{c}{SVM} & \multicolumn{4}{c}{IGMN} & \multicolumn{4}{c}{FIGMN} \\
\hline
breast-cancer & 0.65 & $\pm$ & 0.01 & 0.59 & $\pm$ & 0.02 &           & 0.70 & $\pm$ & 0.02 &          & 0.61 & $\pm$ & 0.04 &           & 0.60 & $\pm$ & 0.00 &          & 0.60 & $\pm$ & 0.00 &         \\
CIFAR-10b & 0.83 & $\pm$ & 0.03 & 0.61 & $\pm$ & 0.19 &           & 0.51 & $\pm$ & 0.03 &          & 0.64 & $\pm$ & 0.14 &           & 0.62 & $\pm$ & 0.20 &          & 0.62 & $\pm$ & 0.20 &         \\
german-credit & 0.79 & $\pm$ & 0.01 & 0.65 & $\pm$ & 0.01 & $\bullet$ & 0.79 & $\pm$ & 0.00 &          & 0.60 & $\pm$ & 0.01 &           & 0.62 & $\pm$ & 0.03 &          & 0.62 & $\pm$ & 0.03 &         \\
pima-diabetes & 0.84 & $\pm$ & 0.00 & 0.64 & $\pm$ & 0.02 &           & 0.81 & $\pm$ & 0.01 &          & 0.73 & $\pm$ & 0.02 &           & 0.69 & $\pm$ & 0.03 &          & 0.69 & $\pm$ & 0.03 &         \\
Glass & 0.81 & $\pm$ & 0.02 & 0.78 & $\pm$ & 0.05 &           & 0.70 & $\pm$ & 0.10 &          & 0.72 & $\pm$ & 0.12 &           & 0.79 & $\pm$ & 0.04 &          & 0.79 & $\pm$ & 0.04 &         \\
ionosphere & 0.95 & $\pm$ & 0.03 & 0.81 & $\pm$ & 0.01 &           & 0.93 & $\pm$ & 0.00 &          & 0.82 & $\pm$ & 0.03 & $\bullet$ & 0.90 & $\pm$ & 0.03 &          & 0.90 & $\pm$ & 0.03 &         \\
iris & 1.00 & $\pm$ & 0.00 & 1.00 & $\pm$ & 0.00 &           & 1.00 & $\pm$ & 0.00 &          & 1.00 & $\pm$ & 0.00 &           & 1.00 & $\pm$ & 0.00 &          & 1.00 & $\pm$ & 0.00 &         \\
labor-neg-data & 0.93 & $\pm$ & 0.01 & 0.79 & $\pm$ & 0.07 &           & 0.94 & $\pm$ & 0.02 &          & 0.94 & $\pm$ & 0.02 &           & 0.91 & $\pm$ & 0.02 &          & 0.91 & $\pm$ & 0.02 &         \\
MNIST & 1.00 & $\pm$ & 0.00 & 0.97 & $\pm$ & 0.03 &           & 0.96 & $\pm$ & 0.00 &          & 0.97 & $\pm$ & 0.04 &           & 0.95 & $\pm$ & 0.05 &          & 0.95 & $\pm$ & 0.05 &         \\
soybean & 1.00 & $\pm$ & 0.00 & 1.00 & $\pm$ & 0.00 &           & 1.00 & $\pm$ & 0.00 &          & 1.00 & $\pm$ & 0.00 &           & 1.00 & $\pm$ & 0.00 &          & 1.00 & $\pm$ & 0.00 &         \\
twospirals & 0.50 & $\pm$ & 0.08 & 0.76 & $\pm$ & 0.02 &           & 0.48 & $\pm$ & 0.00 &          & 0.47 & $\pm$ & 0.03 &           & 0.61 & $\pm$ & 0.12 &          & 0.61 & $\pm$ & 0.12 &         \\
\hline
Average & 0.84 &       &      & 0.78 &       &      &           & 0.80 &       &      &          & 0.77 &       &      &           & 0.79 &       &      &          & 0.79 &       &      &         \\
\hline
\multicolumn{18}{c}{$\bullet$ statistically significant degradation}\\
\end{tabular} \scriptsize \par}
\end{table}

Although not the focus of this work, we could also observe the accuracy of the IGMN algorithm on the tested datasets in comparison to other algorithms available in the Weka software, like Support Vector Machines (SVM) and the state-of-the-art Dropout Neural Networks \cite{hinton2012improving}  (with 50 hidden neurons, 50\% dropout for the hidden layer and 20\% dropout for the input layer; the specific implementation can be found at https://github.com/amten/NeuralNetwork). It was statistically similar to them with $\beta = 0.001$ (now more than one Gaussian component was allowed) and tuning the $\delta$ parameter by 2-fold cross-validation using 3 different values (0.01, 0.1 and 1), but with the additional benefit of a single-scan through data, allowing it to operate on data streams. Results are shown in table \ref{accuracy} (note that, for this experiment, a smaller subset of CIFAR-10 was used, in order to compensate for the higher computational requirements of more Gaussian components). 

\section{Conclusion}\label{sec:conclusion}
We have shown how to work directly with precision matrices in the IGMN algorithm, avoiding costly matrix inversions by performing rank-one updates. The determinant computations were also avoided using a similar method, effectively eliminating any source of cubic complexity for the learning algorithm. This resulted in substantial speed ups for high-dimensional datasets, turning the IGMN into a good option for this kind of tasks. The inference operation still has cubic complexity, but we argue that it has a much smaller impact on the total runtime of the algorithm, since the number of outputs is usually much smaller than the number of inputs. This was confirmed for one-dimensional outputs, which require only scalar operations.

In general, we could see that the fast IGMN is a good option for supervised learning, with low runtimes and good accuracy after adjusting the two main meta-parameters $\beta$ and $\delta$. It should be noted that this is achieved with a single-pass through the data, making it also a valid option for data streams.

\nolinenumbers

\bibliography{bibliography.bib}
%
%
%

\end{document}